\documentclass[conference]{IEEEtran}
\IEEEoverridecommandlockouts
\usepackage{cite}
\usepackage{amsmath,amssymb,amsfonts}
\usepackage{algorithmic}
\usepackage{graphicx}
\usepackage{textcomp}
\usepackage{xcolor}
\usepackage[misc]{ifsym}
\usepackage{makecell}
\usepackage{multirow}
\usepackage{threeparttable}
\usepackage{hyperref}
\def\BibTeX{{\rm B\kern-.05em{\sc i\kern-.025em b}\kern-.08em
    T\kern-.1667em\lower.7ex\hbox{E}\kern-.125emX}}
\begin{document}

\title{A Simple Framework Uniting Visual In-context Learning with Masked Image Modeling to Improve Ultrasound Segmentation}
\author{Yuyue Zhou\textsuperscript{1}, Banafshe Felfeliyan\textsuperscript{1}, Shrimanti Ghosh\textsuperscript{1}, Jessica Knight\textsuperscript{1},  Fatima Alves-Pereira\textsuperscript{1}, \\Christopher Keen\textsuperscript{2}, Jessica Küpper\textsuperscript{1,2}, Abhilash Rakkunedeth Hareendranathan\textsuperscript{1} and Jacob L. Jaremko\textsuperscript{1}
\\
\IEEEauthorblockA{\textsuperscript{1} Department of Radiology and Diagnostic Imaging, University of Alberta, Edmonton, Canada \\
\textsuperscript{2} Department of Biomedical Engineering, University of Alberta, Edmonton, Canada\\}
\href{https://github.com/yuyue2uofa/SimICL}{GitHub repository}

}

\maketitle

\begin{abstract}
Conventional deep learning models deal with images one-by-one, requiring costly and time-consuming expert labeling in the field of medical imaging, and domain-specific restriction limits model generalizability. Visual in-context learning (ICL) is a new and exciting area of research in computer vision. Unlike conventional deep learning, ICL emphasizes the model's ability to adapt to new tasks based on given examples quickly. Inspired by MAE-VQGAN, we proposed a new simple visual ICL method called SimICL, combining visual ICL pairing images with masked image modeling (MIM) designed for self-supervised learning. We validated our method on bony structures segmentation in a wrist ultrasound (US) dataset with limited annotations, where the clinical objective was to segment bony structures to help with further fracture detection.  We used a test set containing 3822 images from 18 patients for bony region segmentation. SimICL achieved an remarkably high Dice coeffient (DC) of 0.96 and Jaccard Index (IoU) of 0.92, surpassing state-of-the-art segmentation and visual ICL models (a maximum DC 0.86 and IoU 0.76), with SimICL DC and IoU increasing up to 0.10 and 0.16. This remarkably high agreement with limited manual annotations indicates SimICL could be used for training AI models even on small US datasets. This could dramatically decrease the human expert time required for image labeling compared to conventional approaches, and enhance the real-world use of AI assistance in US image analysis. 
\end{abstract}


\section{Introduction}
Integration of deep learning methods into medical imaging has the potential to improve the precision and efficiency of disease assessment. However, various challenges remain unsolved, particularly in the ultrasound (US) imaging modality as a result of low contrast, high level of noise, and limited availability of annotated datasets, since data labeling by expert clinicians is time-consuming and costly.
Tissue boundaries are often blurred in US leading to misclassification of objects, contributing to poor performance of popular pretrained models like Meta’s Segment Anything Model on US datasets \cite{heComputerVisionBenchmarkSegmentAnything2023}.  

Automatic segmentation of musculoskeletal structures from US images can improve diagnostic accuracy in a wide range of disease and clinical scenarios and facilitate more effective treatment planning. Notably, US can detect defects in bone representing fractures, allowing diagnosis of certain types of fractures by handheld US probes before, or as an alternative to, performing radiographs \cite{knight2D3DUltrasound2023a}. However, interpreting US images for fractures currently requires meticulous training of human experts.  A deep learning method tailored to assist in US image segmentation could make US imaging more accessible to clinicians who are not imaging experts, thereby improving the accuracy of US assessment in their hands. 

Conventional fully supervised deep learning models, typically designed to be task- and domain-specific, often exhibit limited generalizability without a large labeled training dataset \cite{chenImprovingGeneralizabilityConvolutional2020}. Moreover, they are particularly susceptible to overfitting when trained on relatively small labeled datasets \cite{xuCrossDatasetVariabilityProblem2020}, a common scenario in medical imaging. In recent years there have been some attempts to solve these problems using self-supervised pretraining methods \cite{felfeliyanSelfsupervisedRCNNMedicalImage2023}. However, they still require additional fine-tuning for downstream tasks like semantic image segmentation \cite{zhouSelfSupervisedLearningMore2023}.  Meanwhile, In-context learning (ICL) an emerging research paradigm, originating from natural language processing (NLP), adopts a more human-like training method where models are given an example of an input and output pair and asked to generate the output of a new input based on the given example \cite{dongSurveyIncontextLearning2022}. Compared to conventional methods, ICL allows models to adapt to new tasks quickly. 

In computer vision, visual ICL can provide the model with additional information from images. Visual ICL can be mainly divided into 2 categories. (1) Adding a border made up of learnable visual prompts to the edges of images while models remain frozen. \cite{zhangInstructMeMore2024},\cite{bahngExploringVisualPrompts2022}, \cite{wuUnleashingPowerVisual2022} have been shown to improve data distribution shift and enhance model performance. (2) Considering tasks as image inpainting problems, in which the model is shown an example of image input and output, asking the model to paint the output of a new image, have been successful in natural images \cite{zhangWhatMakesGood2023},\cite{sunExploringEffectiveFactors2023}.  \cite{wangImagesSpeakImages2023}, \cite{wangSegGPTSegmentingEverything2023},  \cite{liuUnifyingImageProcessing2023} added random masking to output images and trained the model to restore output images given input images. \cite {barVisualPromptingImage2022} concatenated paired sets of support and query images and segmentations into one single image and added random masking over the concatenated images. \cite {barVisualPromptingImage2022}'s proposed visual ICL approach, tailored for natural images, used the combination of two inpainting models MAE-VQGAN. In this work, during training, the model's objective was to predict visual tokens of the output of a query image based on the given input-output image pairs, which served as support. However, the model performance relied on a pretrained VQGAN decoder to translate the visual tokens generted by VQGAN encoder to output predictions \cite{barVisualPromptingImage2022}, limiting its application to natural images as there is no VQGAN weights pretrained on medical imaging.

Building upon the idea of using masked image inpainting proposed in \cite{barVisualPromptingImage2022}, in this paper we proposed a new visual ICL and masked image modeling (MIM) based simple segmentation framework named SimICL and tested it on a prospectively collected wrist fracture US dataset to address the challenges posed by limited labeled data. The objective of this work is to enhance model accuracy in segmenting US images as query images, leveraging the strengths of the visual ICL. We implemented our method based on the SimMIM \cite{xieSimMIMSimpleFramework2022} framework designed for self-supervised learning (SSL), instead of MAE-VQGAN model. The SimMIM framework added random masking patches to images and used masked and unmasked areas as input to ViT \cite{dosovitskiyImageWorth16x162021}, and computed the reconstructed output with original image using Mean Absolute Error loss (MAE loss). It is more flexible and more generalizable for other future research as it can be easily adapted to CNN models. Furthermore, unlike \cite{wangImagesSpeakImages2023}, \cite{wangSegGPTSegmentingEverything2023} \cite{liuUnifyingImageProcessing2023} which only masked the output image, inspired by MAE-VQGAN we provided the model with an example pair and a query image and randomly masked over them all.

To the best of our knowledge, our study is the first use of visual ICL in US imaging. In our experiments described below, our novel SimICL framework showed substantial improvement compared to state-of-the-art segmentation and visual ICL models. The broad application of MIM, where the two-stage pretraining-finetuning process was no longer required, had the potential to significantly enhance model performance using only one-stage training combined with proper visual ICL.

\section{Methods}

\subsection{Dataset and data preparation}

Data were collected prospectively at Stollery Children’s Hospital emergency department with institutional ethics approval. We obtained US cine sweeps of the dorsal, volar, and radial aspects of the distal metaphysis and first carpal row in children 0-17 years old using a Philips Lumify L5-12 MHz probe. Cine sweeps were converted into a sequence of single frames and manually segmented by a musculoskeletal sonographer. Only frames with distinct and clear views of pathology were used for our study. Data were anonymized and split into training/validation/test sets based on patient study ID. Due to the redundancy of US images, limited annotations and based on our preliminary experiments, only 20\% of training images with labels were selected for training. For visual ICL models, training images were split into a support pool and a query pool randomly. We randomly selected one image from the support pool, paired it with another from the query pool and repeated the process until all training images were paired. On the test set, since the support image may lead to extra information leakage for the query image, we randomly paired each test image with one support image from the validation set. Dataset details can be found in Table \ref{table1}. An image example can be found in Fig \ref{fig1}A.

\begin{table}[t]
\caption{Image and Patient information}

\label{table1}
\begin{center}
\begin{tabular}{|c|c|c|c|c|c|}
\hline
\multirow{2}*{} & \multicolumn{2}{|c|}{Visual ICL models} &  \multicolumn{3}{|c|}{Segmentation models} \\
\cline{2-6}
~ & Train  & \makecell[c]{Test\\query images}  & Train & Validation & Test \\
 \hline
\# of Images  & 1869  & 3822  & 3738 & 4215 & 3822 \\
 \hline
\# of Patients  & 83  & 18  & 83 & 17 & 18 \\
 \hline
\end{tabular}
\end{center}
\end{table}

\subsection{SimICL}
\begin{figure}[b]
\centering
  \includegraphics[scale=0.6]{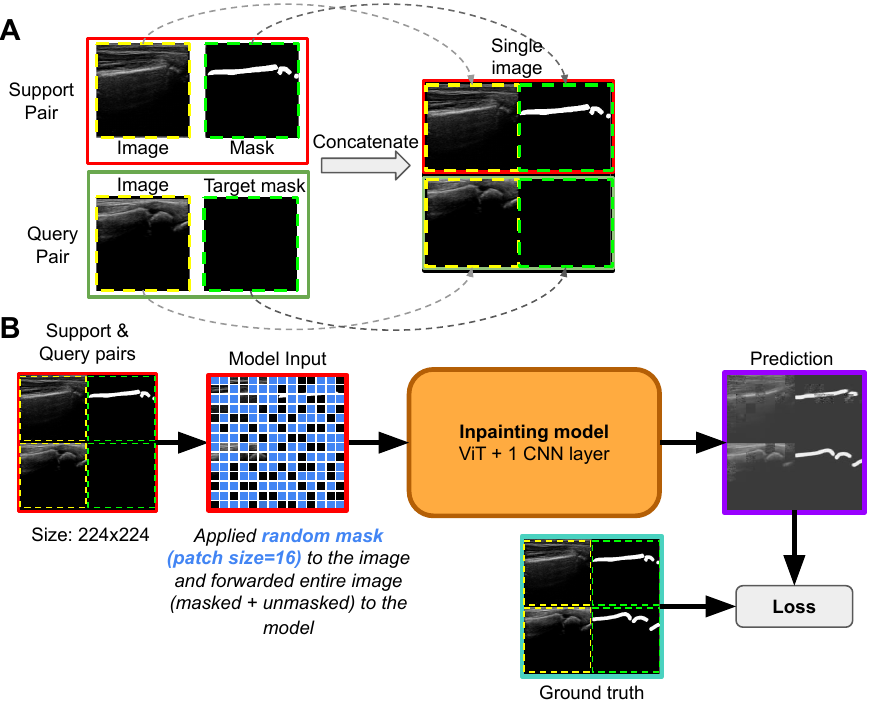}
  \caption{SimICL overview. (A) We constructed a new input image based on one support image/mask pair and one query image. (B) The random mask was added to the image and then the masked image was fed into the model.}
  \label{fig1}
\end{figure}

The concatenated images were resized to 224 x 224. We randomly masked the concatenated input image and fed all the masked and unmasked areas to the model. We set mask patch set 16 based on preliminary experiments and set training mask ratios of 0, 0.3, 0.45, 0.6, and 0.75. The model architecture was kept consistent with the SimMIM framework \cite{xieSimMIMSimpleFramework2022}, with a vanilla ViT encoder with 12 layers and 768 embedding dimensions and a lightweight prediction head using a single convolutional layer. We trained the model from scratch, using MAE loss to compute prediction with ground truth image. Fig \ref{fig1}B shows the overview of our method.

We used AdamW optimizer, and after hyperparameter tuning, we set 0.0005 learning rate and 0.05 weight decay with batch size 64. We trained models for 1200 epochs using V100 GPU.

\subsection{Baseline Models and Evaluation Metrics}
We compared our work with state-of-the-art models: conventional segmentation models  U-Net \cite{ronnebergerUNetConvolutionalNetworks2015d}, nnUNet \cite{isenseeNnUNetSelfconfiguringMethod2021}, self-supervised model SS-RCNN \cite{felfeliyanSelfsupervisedRCNNMedicalImage2023}, and visual ICL models MAE-VQGAN \cite{barVisualPromptingImage2022} and Painter \cite{wangImagesSpeakImages2023}. We did hyperparameter tuning on all five models mentioned above and chose the ones with the best performances for comparison. Models were evaluated on test set images, with Dice coefficient (DC) and Jaccard Index (IoU) as our evaluation criteria. All models were trained using V100 GPU.

\section{Results and Discussion}
\subsection{MAE loss over masked areas or over the entire image is the best training loss setting for SimICL}
Table \ref{table2} shows the quantitative results of different training paradigms on mask patch size 16: (1) training masking ratio (0, 0.3, 0.45, 0.6, 0.75) (2) MAE loss calculation areas (over (a) masked areas (b) all areas (c) areas of support mask + target segmentation and (d) areas of target segmentation). Loss (a)(b) achieved similar and more steady performances on masked images and were more suitable compared to other two. With masked image reconstruction performance evaluation added ((a)(b)), segmentation performance was enhanced compared to loss focusing on segmentation regions only ((c)(d)), indicating that image reconstruction helped the model to learn image representation better, which contributed to segmentation performance improvement. 
\begin{table}[b]
\caption{DC under different training settings}
\label{table2}
\centering
\begin{threeparttable}[b]

\begin{tabular}{|c|c|c|c|c|c|c|}
\hline
 \multicolumn{2}{|c|}{}& (a)\tnote{1}& (b)\tnote{2}& (c)\tnote{3}& (d)\tnote{4} &\makecell[c]{Single image\\+(b)\tnote{2}}\\
\hline
\multirow{5}*{\makecell[c]{Mask\\ratio}} & 0.0  & -  & 0.00 & 0.00 &0.02 & 0.81\\
\cline{2-7}
~ & 0.3  & 0.95  & 0.95 & 0.89 & 0.10 & 0.81\\
\cline{2-7}
 ~ & 0.45  &  \textbf{0.96}  &  \textbf{0.96} & 0.00 & 0.79 & 0.74\\
\cline{2-7}
 ~ & 0.6  &  \textbf{0.96}  & 0.95 & 0.88 & 0.64 & 0.77\\
 \cline{2-7}
 ~ & 0.75  & 0.95  & 0.95 & 0.77 & 0.10 & 0.76\\
 \hline
\end{tabular}
\begin{tablenotes}
\item[1] Loss over masked areas
\item[2] Loss over all areas
\item[3] Loss over segmentation areas
\item[4] Loss over target segmentation area

\end{tablenotes}
\end{threeparttable}

\end{table}

\subsection{Random masks help with query image segmentation}
In Table \ref{table2} loss (b)(c)(d), models performed badly on unmasked images (training masking ratio = 0) while on loss (a)(b) models achieved similar and amazing performances on partially masked images (training masking ratio 0.3-0.75), giving a revelation that proper masking could help with support-query paired image segmentation. We are also surprised to find that even a relatively low masking ratio (0.3) could lead to great segmentation, which is different from MIM in SSL where 0.6-0.8 would be ideal \cite{xieSimMIMSimpleFramework2022} \cite{heMaskedAutoencodersAre2022}. Based on these preliminary experiments, we stuck to adding random masks with MAE loss over (a) masked areas for the following SimICL experiments. 

\subsection{Support pair helps with query image segmentation}

We explored the SimMIM framework on the conventional segmentation, with one single image as input and its corresponding segmentation as the target. We put a random mask with a training ratio of 0.3-0.75 on the input image. All the other settings including model architecture were kept the same as our SimICL. As shown in the final column “single image+(b) loss over all areas” of Table \ref{table2}, the model performed worse compared to those adding support pairs (a)(b), which further demonstrated that the success of SimICL segmentation was based on the given support image-segmentation pairs more than query images themselves.

\subsection{Comparison with other methods}
We compared our work (training mask ratio 0.6, MAE loss (a) over masked regions) with conventional segmentation models U-Net, nnUNet, mask R-CNN, and two masked visual ICL models MAE-VQGAN and Painter. We trained models for different epochs based on their intrinsic characteristics and best performance, for example, we noticed that MAE-VQGAN deteriorated after 100 epochs so we only present the results of 100 epochs here. Besides, for the conventional segmentation network we used different input size and the training time could be different. In Painter \cite{wangImagesSpeakImages2023}, they concatenated two images into one image, and their corresponding two target segmentations into one segmentation and added a random mask only on target segmentations. They fed the concatenated images and segmentation to the model.  Our work concatenated a support pair and a query image into one single image and added a random mask over the newly concatenated image. 

\begin{figure}[t]
\centering
  \includegraphics[scale=0.35]{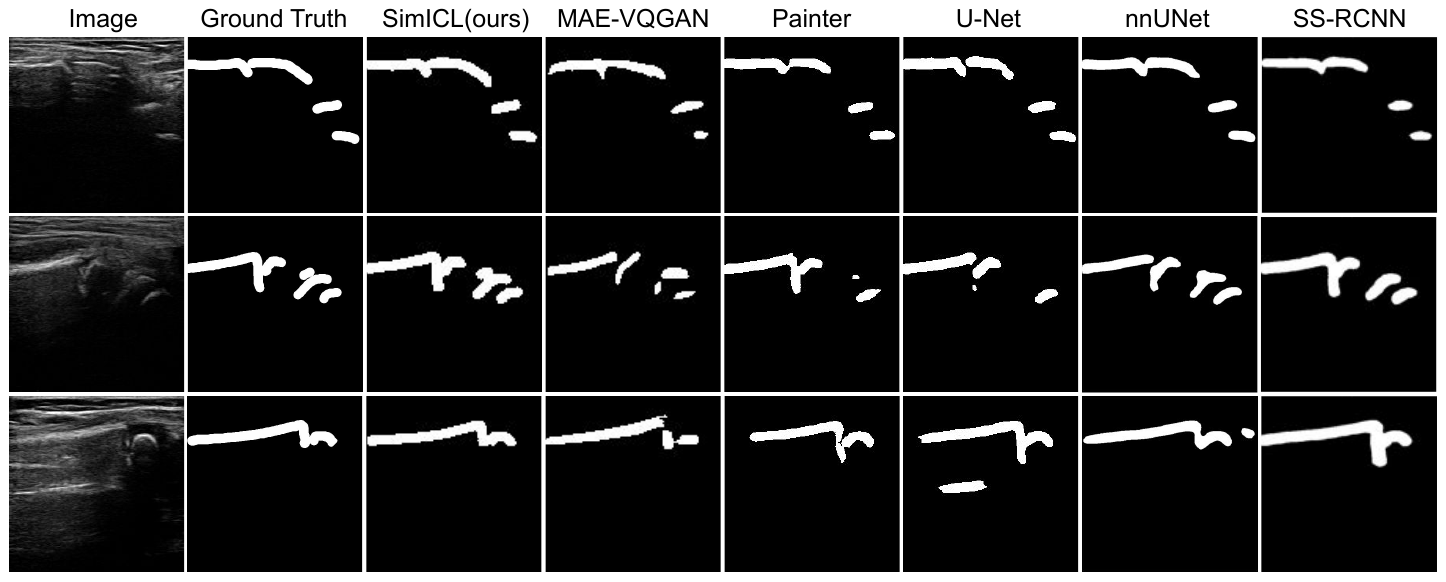}
  \caption{Segmentation prediction on test images.}
  \label{fig2}
\end{figure}

Table \ref{table3} and Figure \ref{fig2} show the quantitative and visualization results of these 6 models. We were not surprised to find that MAE-VQGAN achieved a relativly worse performance compared to other models, as there was no way to update VQGAN decoder weights during training process, and it merely relied on the pretrained weights while there was no  weights pretrained on medical imaging. Besides, the image inpainting models MAE-VQGAN and Painter, designed for multiple tasks including segmentation and edge detection, were slightly inferior than conventional segmentation networks, since the latters were specially designed for image segmentation. 
Our work SimICL achieved significantly better performance compared to the other methods both quantitatively and visually, with DC and IoU increasing by at least 0.10 and 0.16 respectively. SimICL acquired a prediction almost identical to the ground truth segmentation, while the other methods missed bony regions or wrongly segmented the muscle areas. Our model also ran significantly faster than visual ICL model Painter. Those results further demonstrated the strong improvement of our SimICL work compared to other models.

Despite the substantially better performance of SimICL, it did have a drawback with regards to image size limitation: concatenation of four images into one constrains the original image size to 112 pixels per side, leading to relatively rough edge segmentation.  This tool would not be well suited to tiny objects; fortunately, most bones and human anatomic structures are sizeable.  In this work, we deployed SimICL with ViT to one dataset with a single segmentation task. In the future, we will focus on determining how to enlarge the input image size while keeping the model computing costs at a minimum, and apply it to multiple datasets with different tasks and different models to validate its generalization ability.
\begin{table}[t]
\caption{Quantitative results of SimICL and other methods}

\label{table3}
\begin{center}
\begin{tabular}{|c|c|c|c|c|c|c|}
\hline
 & {\makecell[c]{Sim-\\ICL\\(ours)}} & {\makecell[c]{MAE-\\VQGAN\\\cite{barVisualPromptingImage2022}}} & {\makecell[c]{Painter\\\cite{wangImagesSpeakImages2023}}} & {\makecell[c]{U-Net\\\cite{ronnebergerUNetConvolutionalNetworks2015d}}} & {\makecell[c]{nnUNet\\\cite{isenseeNnUNetSelfconfiguringMethod2021}}} & {\makecell[c]{SS-\\RCNN\\\cite{felfeliyanSelfsupervisedRCNNMedicalImage2023}}}\\
\hline
DC & \textbf{0.96}  & 0.78& 0.84 & 0.86 &0.86 & 0.85\\
\hline
IoU  & \textbf{0.92}  & 0.65& 0.73 & 0.75 & 0.76 & 0.74\\
\hline
{\makecell[c]{Train-\\ing\\time\tnote{1}}} & {\makecell[c]{21h\\1200\\epochs}}
 &  {\makecell[c]{2h\\100\\epochs}} & {\makecell[c]{107h\\1200\\epochs}}  & {\makecell[c]{11h\\200\\epochs}}  & {\makecell[c]{12h\\100\\epochs}}  &{\makecell[c]{13h\\20\\epochs}} \\
 \hline
\end{tabular}

\end{center}
\end{table}

\section{Conclusions}
We proposed a new segmentation technique named SimICL combining paired visual ICL images and SSL MIM framework SimMIM, which was well-suited to anatomic structure segmentation on US images. We used SimICL to perform successful experiments detecting bones on wrist US images with limited data labeling. With concatenated support examples and random masking, the model learned the relationship between the support examples and the query images, allowing the precise segmentation of wrist US bony regions, surpassing state-of-the-art segmentation models and visual ICL models, achieving an extremely high DC of 0.96 and IoU of 0.92. SimICL’s robustness on small US datasets could decrease the medical expert labeling time to a large extent. Our innovative SimICL method shows a promising future in visual ICL, and the broad application of MIM.

\section*{Acknowledgment}

Dr. Jacob L. Jaremko is supported by the AHS Chair in Diagnostic Imaging and Canada CIFAR AI Chair, and his academic time is made available by Medical Imaging Consultants (MIC), Edmonton, Canada. We acknowledge the support of TD Ready Grant, Alberta Innovates for financial support, the Alberta Emergency Strategic Clinical Network for clinical scanning, and Compute Canada in providing us with computational resources including high-power GPU that were used for training and testing our deep learning models.

\bibliographystyle{IEEEtran}
\bibliography{EMBC}

\end{document}